\documentclass[sigconf]{acmart}

\AtBeginDocument{%
  }

\setcopyright{acmlicensed}
\copyrightyear{2024}
\acmYear{2024}
\acmDOI{}

\acmConference[HealthRecySys '24]{the 6th International Workshop on Health Recommender Systems co-located with the 18th ACM Conference on Recommender Systems}{October 18,
  2024}{Bari, Italy}
\acmISBN{}

\begin{document}

\title{
Improving the prediction of individual engagement in recommendations using cognitive models}

\author{Roderick Seow}
\email{yseow@andrew.cmu.edu}

\affiliation{%
  \institution{Carnegie Mellon University}
  \city{Pittsburgh}
  \state{Pennsylvania}
  \country{USA}
}

\author{Yunfan Zhao}
\email{yunfanzhao@fas.harvard.edu}
\affiliation{%
  \institution{Harvard University}
  \city{Boston}
  \state{Massachusetts}
  \country{USA}
}

\author{Duncan Wood}
\email{djwood@andrew.cmu.edu}
\affiliation{%
  \institution{Carnegie Mellon University}
  \city{Pittsburgh}
  \state{Pennsylvania}
  \country{USA}
}

\author{Milind Tambe}
\email{milind_tambe@harvard.edu}
\affiliation{%
  \institution{Harvard University}
  \city{Boston}
  \state{Massachusetts}
  \country{USA}
}

  \author{Cleotilde Gonzalez}
\email{coty@cmu.edu}
\affiliation{%
  \institution{Carnegie Mellon University}
  \city{Pittsburgh}
  \state{Pennsylvania}
  \country{USA}
}

\renewcommand{\shortauthors}{Seow et al.}


\begin{abstract}
For public health programs with limited resources, the ability to predict how behaviors change over time and in response to interventions is crucial for deciding when and to whom interventions should be allocated. Using data from a real-world maternal health program, we demonstrate how a cognitive model based on Instance-Based Learning (IBL) Theory can augment existing purely computational approaches. Our findings show that, compared to general time-series forecasters (e.g., LSTMs), IBL models, which reflect human decision-making processes, better predict how individuals' behaviors change over time (transition-consistency) and in response to receiving an intervention (intervention-sensitivity). We further show that IBL parameters capture the individual differences in transition-consistency and intervention-sensitivity and that other time series models can use these individual-level IBL parameters to improve their training efficiency.
\end{abstract}





\maketitle

\section{Introduction}

Public health programs play an essential role in improving the health outcomes of individuals and communities, often through education and subsequent behavioral change. Some health programs interact with their intended beneficiaries in a broad and infrequent manner. For example, a campaign about the health risks of smoking may address a general population of smokers through scattered advertisements in the media \cite{nogueira2018impact}. Others rely on repeated direct interactions with their intended beneficiaries. For example, maternal health programs that send automated messages about exercise and nutrition to enrolled expectant mothers \cite{lalan2024improving}. In this case, it is crucial that mothers remain engaged for the duration of the program or as long as possible to receive the maximum benefit. Unfortunately, many such programs face high levels of disengagement and dropout, which severely limit their effectiveness \cite{amagai2022challenges}.

To reduce dropout, programs can provide interventions designed to increase beneficiary engagement. For example, program staff can make personalized service calls to beneficiaries at risk of dropping out to address concerns about declining participation. However, programs with limited staff and resources cannot provide interventions to all their beneficiaries. Instead, they must choose a subset of beneficiaries to receive the intervention.  

Recent approaches to this challenge have adopted the Restless Multi-Armed Bandit (RMAB) resource allocation framework \cite{killian2023robust}. Classically, the RMAB framework models the engagement dynamics of each beneficiary (i.e., the transition between discrete engagement and disengagement states) as a Markov Decision Process (MDP). A centralized planner decides which beneficiaries to intervene on at each time point based on the dynamics learned from the MDPs. Importantly, beneficiaries can change their engagement state even without an intervention. The best-performing approach to this framework is the Whittle index \cite{whittle1988restless}, which ranks beneficiaries by their likelihood of becoming and remaining engaged after getting an intervention.

This framework assumes that beneficiaries' dynamics---how their level of engagement changes over time---can be modeled as an MDP. However, recent work has demonstrated that these dynamics are often non-Markovian; that is, a beneficiary's transition between being engaged and disengaged with the program (their state) depends on their individual history \cite{danassis2023limited}. As a substitute, Danassis and colleagues (2023) developed the Time-series Arm Ranking Index (TARI), which allows non-Markovian dynamics and continuous engagement levels within the RMAB framework. The TARI approach models beneficiaries using time-series forecasters such as LSTMs \cite{hochreiter1997long} or transformers \cite{vaswani2017attention} instead of MDPs. Using these learned time-series models, TARI computes an index that captures the relative engagement benefit of intervening on a beneficiary by comparing their expected future engagement with and without an intervention at the current time point and then ranks beneficiaries according to the computed indices.

Within the TARI approach to the RMAB framework, we propose modeling individual beneficiaries with computational cognitive models instead of LSTMs. Specifically, we create a model based on Instance-Based Learning Theory (IBLT) \cite{gonzalez2003instance} to represent each beneficiary's time series activities. We expect using an IBL model will convey two advantages over general time-series forecasters (TSFs). 

First, IBL models are personalized to the individual instead of trained across the entire dataset, allowing the models to capture individual differences in behavioral dynamics (i.e., individual-level differences in the response to interventions and general engagement patterns). Because TSFs need large amounts of data to fine-tune their many parameters and there is relatively sparse data available per beneficiary, existing approaches train a single model using the combined data from all available beneficiaries. For instance, Danassis and colleagues (2023) create a set of fixed-length vectors from each beneficiary's trajectory. Each vector represents the intervention history and engagement levels of seven consecutive time steps. Then an LSTM trains on these vectors to predict the engagement level at the next time step given the previous seven \cite{danassis2023limited}. In contrast, an IBL model has enough cognitively grounded structure that it can be applied to an individual with sparse data \cite{bugbee2022making}. This is because IBL theory makes strong assumptions about the cognition of human decision making derived from prior research. In particular, IBL theory accounts for memory effects (i.e., the more recent a data point is, the more it contributes to the prediction) and similarity effects (i.e., engagement levels that are observed under similar contexts should be similar to each other) that general TSFs do not.

Second, the parameters of IBL cognitive models are grounded in psychological constructs such as memory and attention, making them more interpretable than the parameters of general LSTMs. This advantage allows cognitive models to shed light on the individual differences that drive beneficiaries' behaviors. We demonstrate how these individual differences can be used to augment existing approaches that rely primarily on general LSTMs.

\subsection{Our Contributions}

\begin{enumerate}
    \item{Using real-world data from a maternal health program, we show that cognitive models more accurately capture the temporal dynamics of individual behaviors and predict the behaviors of individual beneficiaries than LSTMs.}
    \item {We showcase that personalized cognitive models could reveal individual characteristics through weights fitting. We illustrate how to cluster individuals using learned characteristics.}
    \item {We show clustering based on learned individual characteristics could guide and improve the performance of time series models. Specifically, we show that LSTMs trained 
    in clusters better predict behaviors within those clusters compared to LSTMs trained on random samples of beneficiaries.}
\end{enumerate}

\subsection{Background}

\subsubsection{Time-series Arm Ranking Index (TARI)}

In the context of multiple potential beneficiaries treated as arms in an RMAB, each arm is considered an independent time series \cite{danassis2023limited}. We train a model to predict the next state $s_{t+1}$ based on three pieces of information:
\begin{itemize}
\item A historical record (length h) of the past states and actions of the arm (denoted by $\{(s_i, a_i)\}_{i<t}^h$).
\item The current state (e.g., engagement level) of the arm $s_t$.
\item A potential action $a_t$, where 1 represents an intervention and 0 represents no intervention.
\end{itemize}
This training takes place offline. During test time, the model's parameters are fixed, and we use the trained model with iterated multi-step forecasting to generate a long-term forecast of future states $s_{t+1}, s_{t+2}, ..., s_{t+H}$ \cite{taieb2012recursive}. 

For each arm, we use the TSF model to estimate two values:
\begin{itemize}
\item Time to disengagement \textbf{with} intervention $u_n$: The predicted number of timesteps until the arm becomes non-engaged if we intervene at the current timestep (t) and never again.
\item Time to disengagement \textbf{without} intervention $v_n$: The predicted number of timesteps until the arm becomes non-engaged with no interventions at any point.
\end{itemize}
The TARI index for an arm is the ratio of these two times. A higher TARI index suggests that intervening now would be more beneficial. Finally, similar to the Whittle index, we choose to act on the $k$ arm with the highest TARI index at each timestep:

$$\text{TARI}(n) = \frac{u_n}{v_n}.$$

\subsubsection{Instance-Based Learning Theory}

Instance-Based Learning Theory (IBLT) is a cognitive theory of dynamic decision making, grounded in human learning and memory mechanisms. It has successfully accounted for human behavior in a variety of contexts, ranging from abstract repeated-choice tasks to more real-world search and choice \cite{nguyen2022theory} and continuous control tasks \cite{gonzalez2003instance}. In most of these contexts, human participants are required to make a series of decisions while exposed to changing environments. 

The essence of IBLT is storing past decisions in the form of an instance. An instance has three parts: the context, the choice made, and the associated utility. For example, the cloudiness of the sky could be the context, the choice to bring an umbrella could be the choice, and the utility could be the inconvenience of bringing an umbrella when it does not rain (supposing it did not rain in this instance).

IBL agents compute the expected utility of a choice by considering past instances where they made that choice, weighted by the contextual similarity to the present. For example, an IBL agent would compute the expected utility of bringing an umbrella by considering its past experiences with an umbrella, most strongly considering the experiences with the most similar cloudiness to the current moment.

More precisely, the expected value of a choice is the weighted average of the utilities from all the past instances where that choice was made weighted by the retrieval probability (i.e., the probability of remembering that instance) to generate estimates of the choice's expected utility called a blended value. The agent will deterministically make the available choice with the highest computed blended value (expected utility). The utilities of instances in memory are the actual outcome the agent observed when it chooses, regardless of what it predicted. After choosing, a new instance is added to the agent's memory, reflecting what happened. If the exact combination of context, choice, and utility has appeared before, that instance instead has its frequency and recency information updated. 

The contribution of an instance's utility to a choice's expected value depends on the instance's memory activation (the salience of that instance given the current context). The activation of an instance (\(A_i\)) reflects how readily it comes to mind, which in turn depends on the frequency, recency, and similarity factors according to the following equation as proposed by the ACT-R cognitive architecture \cite{anderson2014atomic}.

\begin{equation}
    A_i = ln(\sum_j(t-t_{i,j}')^{-d}) + \mu\sum_k({w_k(Sim(s_{i,k}, s_{t,k}) - 1}) + \sigma\xi
    \label{eq:activation}
\end{equation}

Three expressions additively determine an instance's activation (salience in memory). The first expression is the contribution of frequency and recency where \(t\) is the current timestep, \(t_{i,j}\) is the timestep of the \(j\)th appearance of instance \(i\), and \(d\) is a decay parameter. The second expression is the contribution of similarity where \(\mu\) is a scaling parameter on the overall impact of similarity to activation, \(w_k\) is the importance (or weight) of an attribute in the similarity calculations, \(Sim\) is a similarity function that returns the degree of similarity for a particular attribute \(k\) between instance \(i\) (\(s_{i,k}\)) and the current context (\(s_{t,k}\)). The third expression adds some randomness to the memory retrieval process where \(\sigma\xi\) is a scaled noise distribution. The activation of an instance is then converted into a retrieval probability using the Boltzmann softmax function, where \(\tau\) is the temperature parameter:

\begin{equation}
    P_{i} = \frac{e^{A_i / \tau}}{\sum_{j}e^{A_j / \tau}}
\end{equation}

 The IBL model computes the expected utility of a choice \((a)\) from the utility \(u_i\) of each retrieved instance \(i\) weighted by their respective retrieval probability \(P_i\) \cite{gonzalez2003instance,gonzalez2011instance}:

\begin{equation}
    V(a) = \sum_{i} P_i u_i
\end{equation}

The agent then makes the choice with the highest expected utility, \(V(a)\).

\section{Proposed Approach}

We model each beneficiary with their own individual-level IBL model. We do not have the IBL model make a choice. Instead, we use the IBL's expected utility calculation to predict the engagement level in the next period (see Figure~\ref{fig:sim-setup}a).  Because we only use the model's blended value, the choice part of the instance is always held constant. The utility of an instance is a beneficiary's engagement level (thus, the expected utilities are weighted averages of past engagement levels then used to predict future engagement levels). The context has two components derived from a beneficiary's history: (1) the engagement level of the previous period and (2) the number of timesteps since the last intervention. The IBL model determines the similarity between the current context and each instance in memory according to these two attributes, and these similarities are factored into an instance's activation according to Equation ~\ref{eq:activation}. 

The IBL models are trained through model tracing: recording each timestep of an individual beneficiary as an instance with the relevant context (previous engagement level and time since last intervention) and utility (engagement level). Thus, the model ``traces'' the beneficiary it is modeling by giving itself memories as if it were the beneficiary that produced the data. 

As shown in Equation ~\ref{eq:activation}, the similarity of two instances depends not only on the similarity between attribute values, but also on the weight or influence of the attribute ($w$ in Equation ~\ref{eq:activation}). Psychologically, an attribute weight represents how relevant the agent considers an attribute to make similarity judgments between two contexts. To capture potential individual differences between beneficiaries, we personalized an IBL model for a specific beneficiary by finding the combination of attribute weight values that resulted in the smallest training prediction error.

Given the predictions of engagement level, we then follow the TARI algorithm and generate a ranked list of beneficiaries according to the predicted relative benefit in the improvement in engagement each beneficiary receives from an intervention versus not receiving one.

\section{Simulation Setup}

\subsection{Data}

We tested our approach on data from a maternal healthcare program operated by an NGO called ARMMAN \cite{ARMMAN}. This program sends automated messages about maternal and infant care to expectant and recent mothers in vulnerable communities. However, over the many weeks of pregnancy and postnatal care, many mothers stop engaging with the automated messages. ARMMAN has a limited capacity to intervene: Program staff can call mothers individually, allowing them to hear from a real person and ask questions, hopefully increasing their future engagement. 

ARMMAN collected the data in 2022 from 12,000 mothers over 40 weeks. The data have two recorded values per mother per week: the amount of time she spent listening to the automated health message (engagement level, recorded as a number in $[0,1]$) and if she received a call from ARMMAN staff (if she received an intervention, recorded as a binary value). The data also contain mothers' demographic information; however, previous work found little or no prediction benefit in incorporating demographic variables as additional features for time-series forecasters \cite{danassis2023limited}. Thus, we do not consider them in this approach.

For an IBL model to predict an intervention's effect on a mother's engagement level, it needs to be trained on a trajectory that includes at least one intervention. However, of the 12,000 mothers, only 5,400 mothers received at least one intervention early on in their program tenure. We also assume that mothers receive an intervention (or the equivalent of an intervention) upon their enrollment in the program. Because the time needed to train all the IBL models scales with the number of mothers, we explored our approach with a subset of 210 mothers out of the 5,400 mothers who received at least one intervention in their actual trajectories.

\begin{figure*}[ht!]
    \centering
    \includegraphics[width=\linewidth]{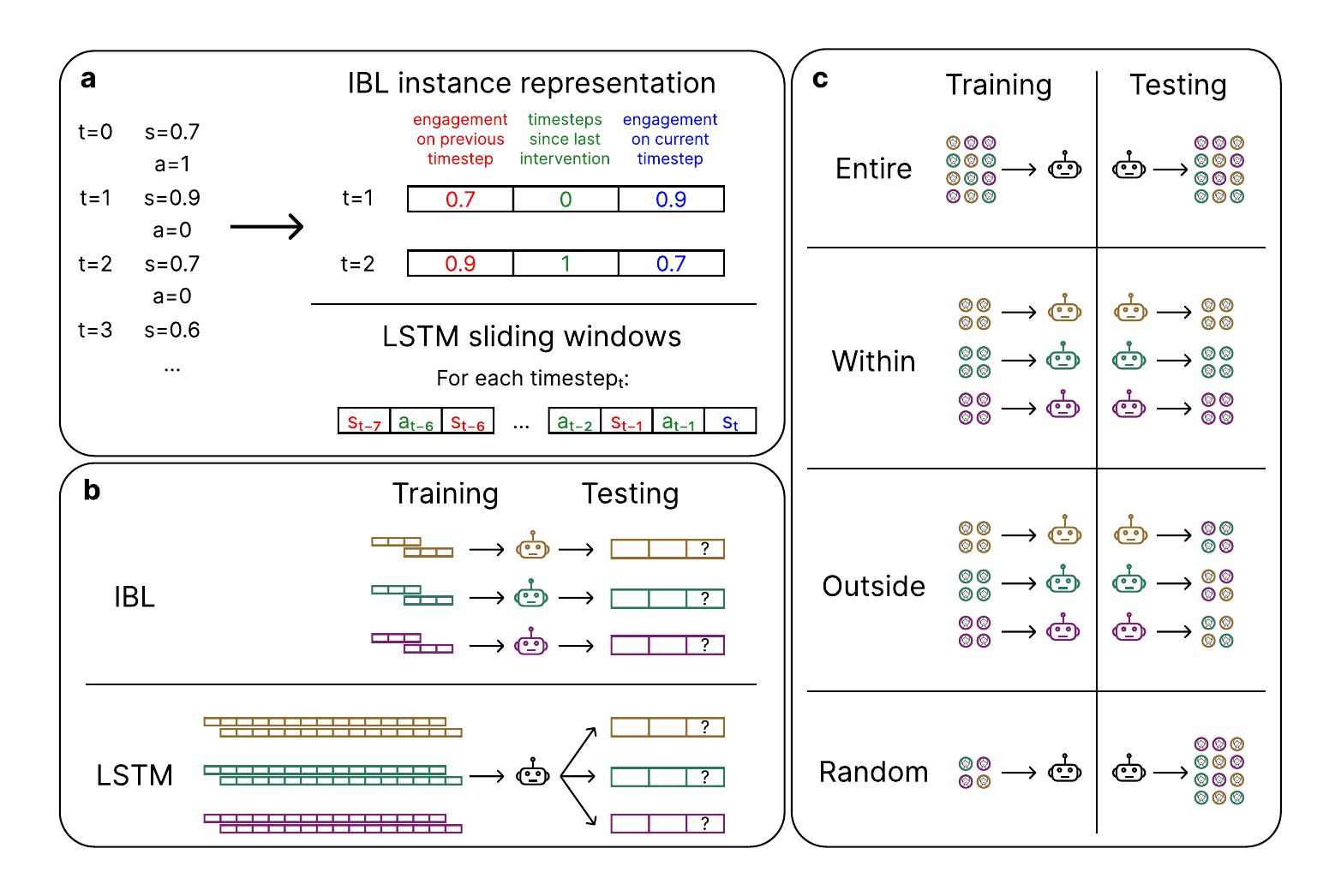}
    \caption{\textbf{a)} Transforming a beneficiary's trajectory into IBL instances and LSTM sliding windows. Each IBL instance represents the engagement context at a given timestep and consists of three parts: the engagement level on the immediately prior timestep, the number of timesteps since the last intervention, and the engagement level on the current timestep. In contrast, the LSTM data was constructed by segmenting trajectories into sliding windows of length 7. \textbf{b)} Training / testing setup for comparing IBL and LSTM methods. Different colors indicate data from individual beneficiaries. Under the IBL procedure, one IBL model is trained per beneficiary. Each trained IBL model is then used to predict their respective beneficiaries during testing. Under the LSTM procedure, one LSTM is trained using data from all beneficiaries. The trained LSTM model is then used to make predictions for all beneficiaries during testing. \textbf{c)} LSTM training / testing setups under various IBL-clusters-related conditions. Different colors indicate different cluster types. Note that the LSTM in the Random condition is trained on a random subset of one-third of the training data as compared to the other training conditions.}
    \Description{A line graph of the prediction error of the two model types - IBL and LSTM - over time.}
    \label{fig:sim-setup}
\end{figure*}

\subsection{Simulation Procedure}

We designated the first 25 weeks of each mother's trajectory as training data. Following the approach of previous studies, we reconstructed the training data into sliding windows of 7 consecutive time steps to train the LSTM model. Then, we trained an LSTM model on the entire training dataset (see Figure ~\ref{fig:sim-setup}b for a visual description of the differences between the IBL and LSTM training / testing procedure). 

Each IBL model (one for each mother) traced a mother's trajectory through each time step up to week 25. To find the best-fitting profile of attribute weights per mother, a grid search looked for the combination that minimized the weighted sum of one-step prediction errors (loss). The grid iterates all possible combinations of parameter values in the range (0, 0) to (5, 5) in intervals of 0.5. 

An IBL model should improve its predictive accuracy as it accumulates more instances and becomes more ``familiar'' with the individual it's modeling. So, we weight earlier prediction errors (e.g., at weeks 1 and 2) less than later ones (e.g., at weeks 24 and 25) according to the following equations:

\begin{equation}
    loss = \sum_t{q_t * SE_t}
\end{equation}

\begin{equation}
    q_t = \frac{e^{t / 10}}{\sum_i{e^{t_i / 10}}}
\end{equation}

where $SE_t$ is the squared prediction error and $q_t$ is the normalized weight factor for the timestep $t$. 

The trained models were then tasked with generating predictions for the next 14 weeks. For the one-step prediction task, the models iteratively generated predictions for the next time step (i.e., for week 26, then for week 27, then for week 28, etc.). Consistent with the fact that the sampled mothers did not receive additional interventions beyond the first 14 weeks of their tenure, we simulated these predictions without additional simulated interventions, which allowed us to compare the models' engagement predictions with the ground truth data.

We compared the IBL-TARI and LSTM-TARI policies with three other baseline policies: (i) random allocation, where each beneficiary has the same probability of receiving an intervention, (ii) round-robin allocation, where a beneficiary that received an intervention will not receive another until all other beneficiaries have also received an intervention, and (iii) no intervention, where no interventions are delivered. Consistent with ARMMAN's limited resources, all policies are allowed a 3\% budget (except the no intervention policy), which corresponds to allocating interventions to 6 beneficiaries.

To simulate the effect of an intervention on a beneficiary's engagement level when they did not receive one in reality, it was necessary to develop a counterfactual generator. We followed the approach adopted in previous studies and trained a separate LSTM model on trajectories from the entire dataset (approximately 12000 beneficiaries), which is far larger than the subset used for our policy simulations. This counterfactual generator only operates when a mother's simulated trajectory deviates from her ground-truth trajectory (i.e., received a simulated intervention when she did not receive one in reality or not receiving a simulated intervention when she did receive one in reality).

\section{Results}

\subsection{Predicting next-step engagement levels}

\begin{figure}[ht!]
    \centering
    \includegraphics[width=\linewidth]{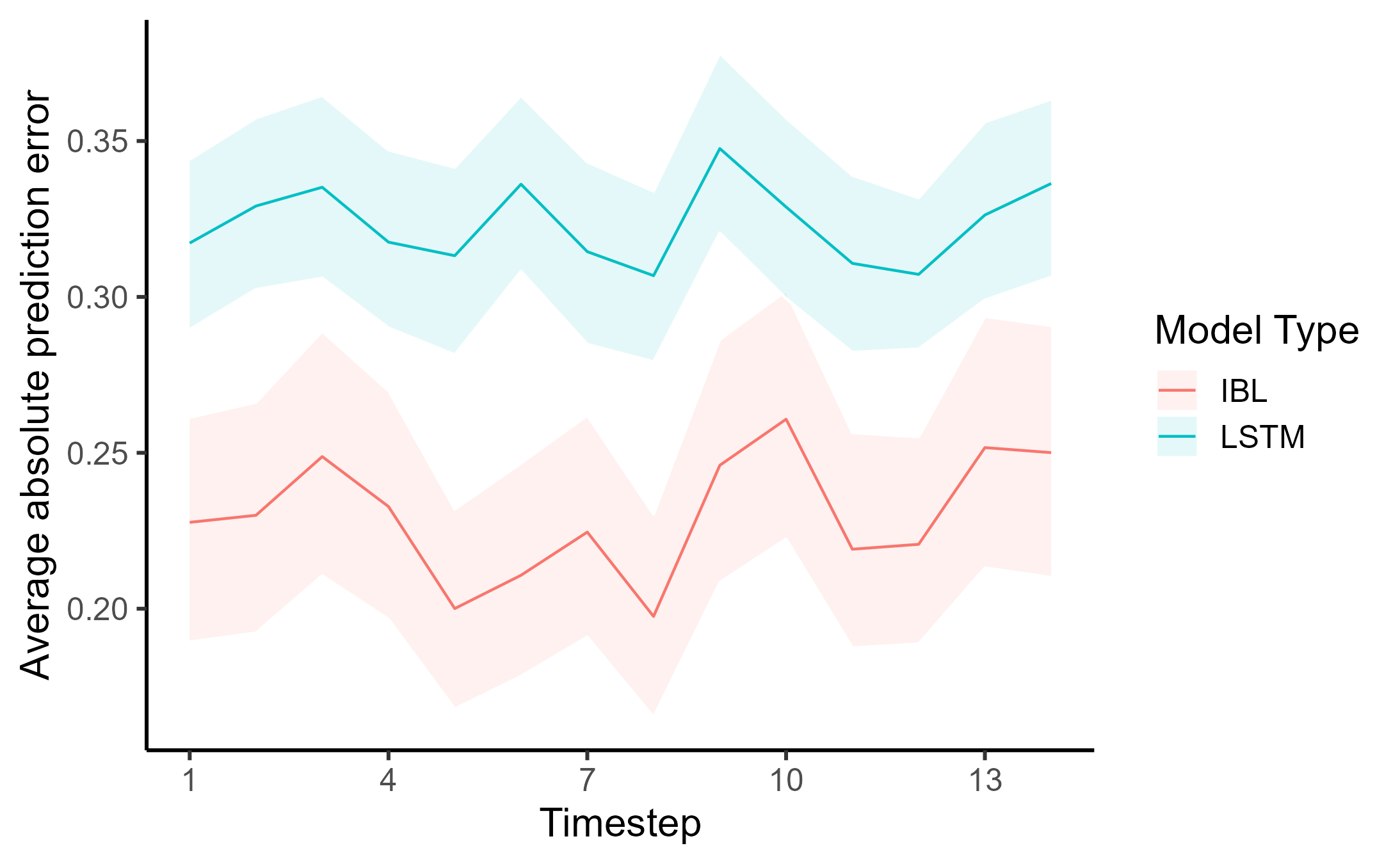}
    \caption{Next-step prediction error per testing time step (\underline{smaller is better}). \textit{N} = 210. The personalized IBL models consistently achieves lower average prediction error (0.23 across all time steps) as compared to the LSTM model (0.32), which translates to a 29\% reduction in error.}
    \Description{A line graph of the prediction error of the two model types - IBL and LSTM - over time.}
    \label{fig:one-step}
\end{figure}

We begin by presenting the results directly comparing the performance of the personalized IBL models and the LSTM model on the next-step prediction task. Figure~\ref{fig:one-step} shows the prediction error of the two approaches during each of the 14 weeks in the testing period. Clearly, IBL models consistently achieve lower prediction errors than LSTM. About 10\% less errors on average (0.23 on average) compared to the LSTM model (0.32 on average).

\subsection{Using predictions to inform intervention allocation}

Given this result, we hypothesized that the TARI algorithm would benefit from using IBL models to fill the role of its time-series forecaster. More accurate next-step predictions should lead to a more accurate prediction of the relative benefit of intervening versus not intervening on a mother, and thus allow for a more accurate ranking of mothers. 
Figure~\ref{fig:percent-engaged} (in Appendix) shows the percentage of mothers who have an engagement level of 0.25 and above during each test time step following the various intervention allocation policies. The LSTM-TARI policy outperforms all other policies, particularly later on in the testing period. The average percentage of engaged mothers across testing timesteps for the various policies are as follows: IBL-TARI ($55.58\%$); Round-robin ($55.65\%$); Random ($54.29\%$); LSTM-TARI ($61.05\%$); Control ($46.22\%$). Contrary to our expectations, the IBL-TARI policy does not outperform the LSTM-TARI policy on this performance metric. 


\subsection{Elucidating individual differences with IBL parameters}

Beyond more accurate predictions of engagement, another advantage of the proposed IBL approach over more general TSFs (as exemplified by the LSTM approach) is the ability to identify individual differences in beneficiaries' behavioral characteristics. Because a separate IBL model models each beneficiary, the parameter values of each model provide potential insights into how beneficiaries differ from each other in their engagement and intervention dynamics.

Specifically, the attribute weights of each mother's IBL model trained on data from the first 25 weeks of her tenure can be interpreted as measures of how reliably her engagement changes in response to an intervention (intervention-sensitivity) and how consistently her engagement changes between time steps (transition-consistency).

\begin{figure}[ht!]
    \centering
    \includegraphics[width=\linewidth]{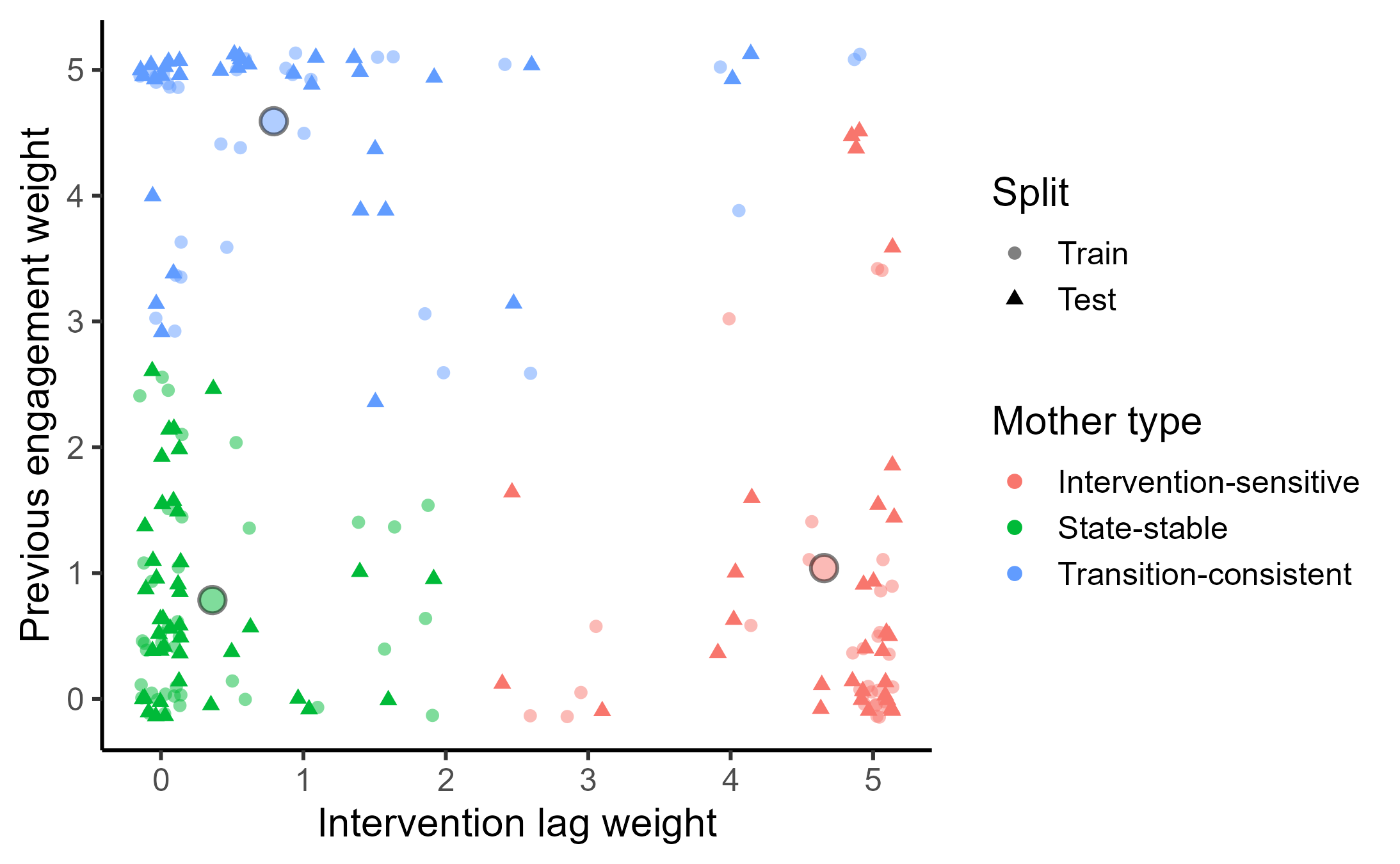}
    \caption{Distribution of attribute weight profiles. Each dot represents the IBL model personalized to a particular beneficiary (mother). Points are slightly jittered to better visualize density. Lighter points indicate training set mothers while darker points indicate testing set mothers.  Beneficiaries are clustered into 3 clusters according to a k-means algorithm. Larger circles outlined in black indicate cluster centroids.}
    \label{fig:weight-dist}
\end{figure}

Figure~\ref{fig:weight-dist} shows the distribution of mothers according to their attributes weight profiles. From the visual inspection of the data, there appear to be three types of mothers. We confirmed that the optimal number of clusters is between 3 to 4 clusters using multiple measures of cluster goodness (silhouette: 4, within-sum-of-squares: 4, gap statistic: 3). For interpretability, subsequent analyzes assume that mothers are partitioned into three clusters. 

Each cluster was labeled according to its defining characteristics. Intervention-sensitive mothers (\(N\) = 62) are those characterized by a high intervention lag weight; given that such a mother is \(t\) time steps away from her most recent intervention, it is likely that her engagement level will be similar to her engagement level when she was \(t\) time steps away from previous interventions. Transition-consistent mothers (\(N\) = 66) are those who are characterized by a high previous engagement weight; Since \(x\) was the engagement level of such a mother in the immediately previous time step, it is likely that her current engagement level will be similar to her engagement level in the time step following a previous state when her engagement level was similar to \(x\). State-stable mothers (\(N\) = 82) are those characterized by relatively low values in both dimensions; given that \(x\) was the level of engagement of such a mother in the immediately preceding time step, its current level of engagement is likely to be similar to \(x\).


\subsubsection{Testing the generalizability of individual differences in IBL attribute weights}


The previous results showed that individual mothers are associated with different weight profiles of the attributes of the IBL instance. However, it remains unclear whether these weights are generally meaningful; that is, if they do capture individual differences that are predictive of behavioral dynamics beyond an IBL modeling context. Specifically, if these attribute weights reflect underlying individual differences in engagement and intervention dynamics, then we would expect other modeling approaches that account for these attribute weights to perform better than those that do not. 

We tested this hypothesis by training and testing LSTM models with various cluster-related samples. Using the cluster assignments derived above, we randomly split each cluster in half to construct a training and a testing set. For all methods, the task is to predict the next-state engagement of the testing set during the testing period (i.e., the last 14 weeks). We examined the following methods (see ~\ref{fig:sim-setup}c for a visual description):
\begin{itemize}
\item{\textbf{Entire}: One LSTM trained on all training mothers; predict for all test mothers.}
\item{\textbf{Within-cluster}: One LSTM trained on each cluster; predict for test mothers in their respective training cluster. For instance, the LSTM trained on the state-stable training set would be tested on the state-stable testing set. We hypothesize that this method should yield the most accurate predictions given the informativeness of the identified IBL clusters.}
\item{\textbf{Outside-cluster}: One LSTM trained on each cluster; predict for test mothers from a different cluster. As there are 3 clusters, we repeat this method to obtain predictions on the other non-training cluster and record the average prediction error of the two simulations. For instance, prediction accuracy for the state-stable testing set is the average of the prediction accuracy of the LSTM model trained on the transition-consistent training set and that of the model trained on the intervention-sensitive training set.}
\item{\textbf{Random}: One LSTM trained on a random subset of one-third of the training set ($N=35$); predict for all test mothers}
\end{itemize}

\begin{figure}[ht!]
    \centering
    \includegraphics[width=\linewidth]{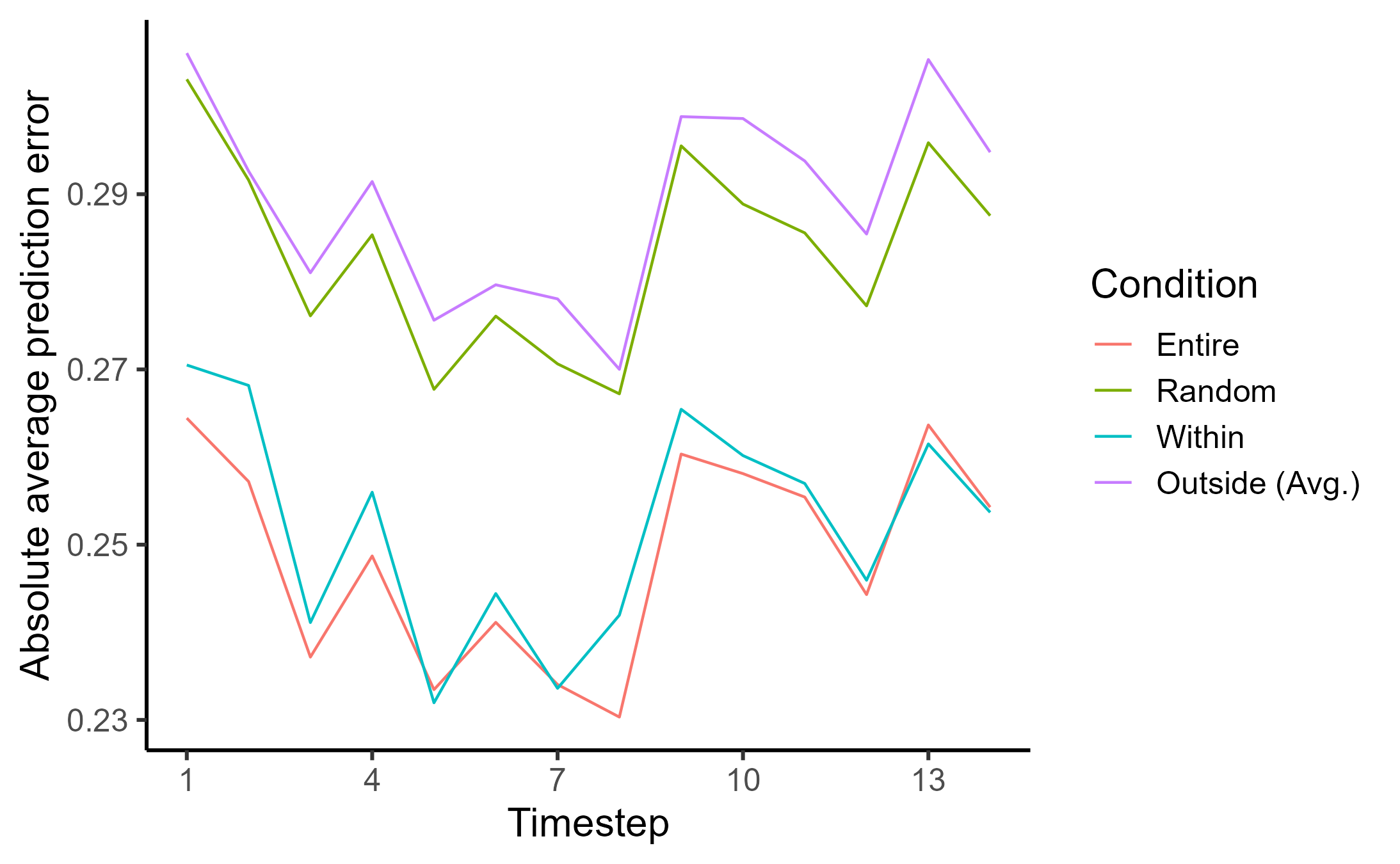}
    \caption{Next-step prediction error per testing time step for the various LSTM training-testing methods. Note that the performance for the Outside-cluster method is an average of the performance of two LSTM models (e.g., for a state-stable mother, the average prediction error of a LSTM trained on intervention-sensitive mothers and a LSTM trained on transition-consistent mothers.}
    \label{fig:lstm-cond-error}
\end{figure}

\begin{figure}[ht!]
    \centering
    \includegraphics[width=\linewidth]{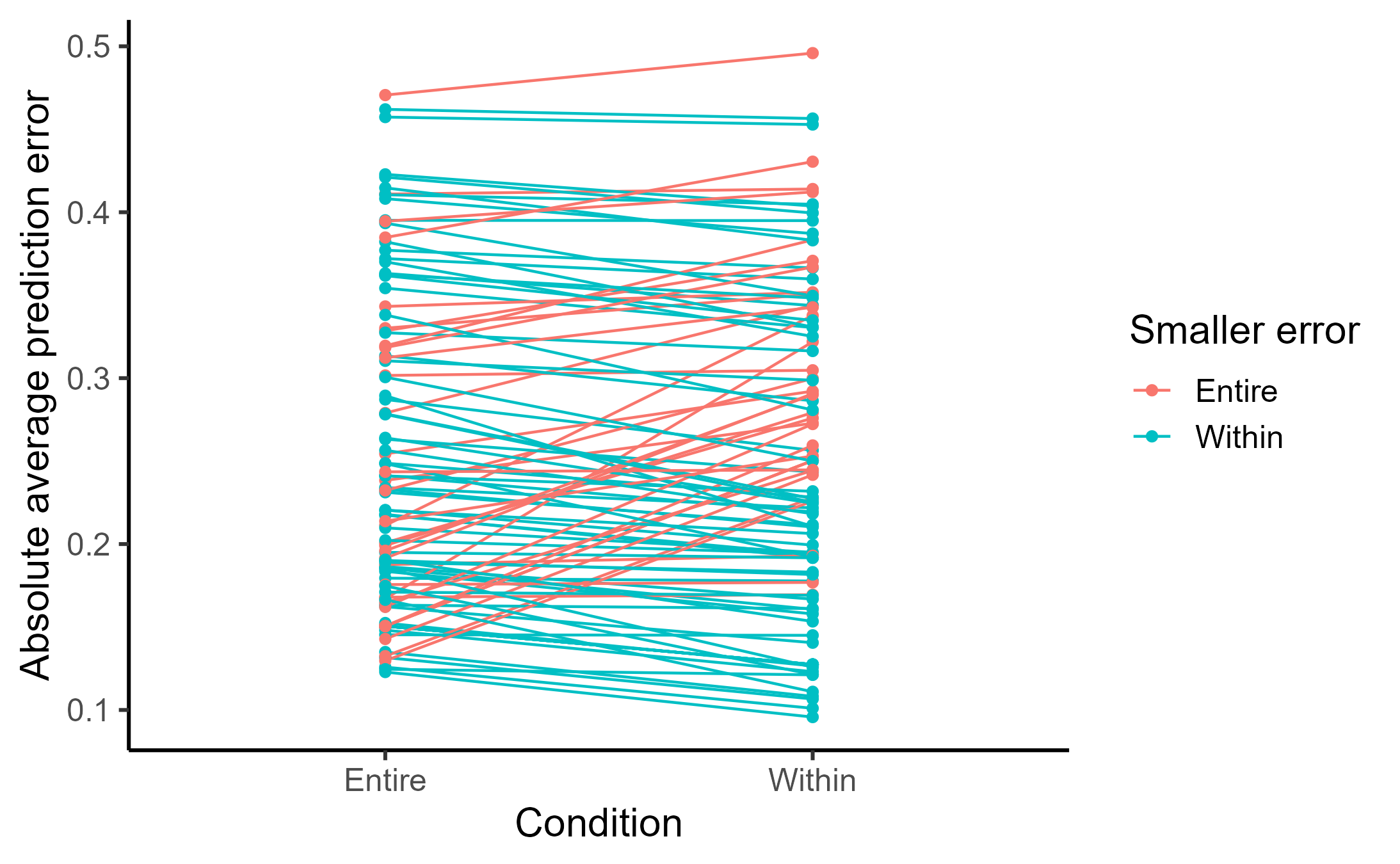}
    \caption{Comparison of prediction errors per testing mother between Entire and Within-cluster methods. Blue lines indicate mothers who are (on average) better predicted by the Within-cluster method whereas red lines indicate those better predicted by the Entire method.}
    \label{fig:lstm-error-counts}
\end{figure}

Figure~\ref{fig:lstm-cond-error} shows the average prediction error for the various methods. In support of the generalizability of the clusters derived from IBL attribute weights, LSTMs that are trained and tested on different clusters (Outside-cluster) and a LSTM trained on a random sample (Random) perform relatively worse compared to the LSTMs that were trained and tested on the same clusters (Within). 

Compared to the Entire method, the Within-cluster method does not appear to provide any additional benefit in average prediction accuracy, which might lead one to favor the simpler approach (i.e., training one vs. three models). However, as Figure~\ref{fig:lstm-error-counts} shows, the Within-cluster method is the more accurate one for two-thirds of the mothers (Within: 70 vs. Entire: 35). 

\begin{figure}[ht!]
    \centering
    \includegraphics[width=\linewidth]{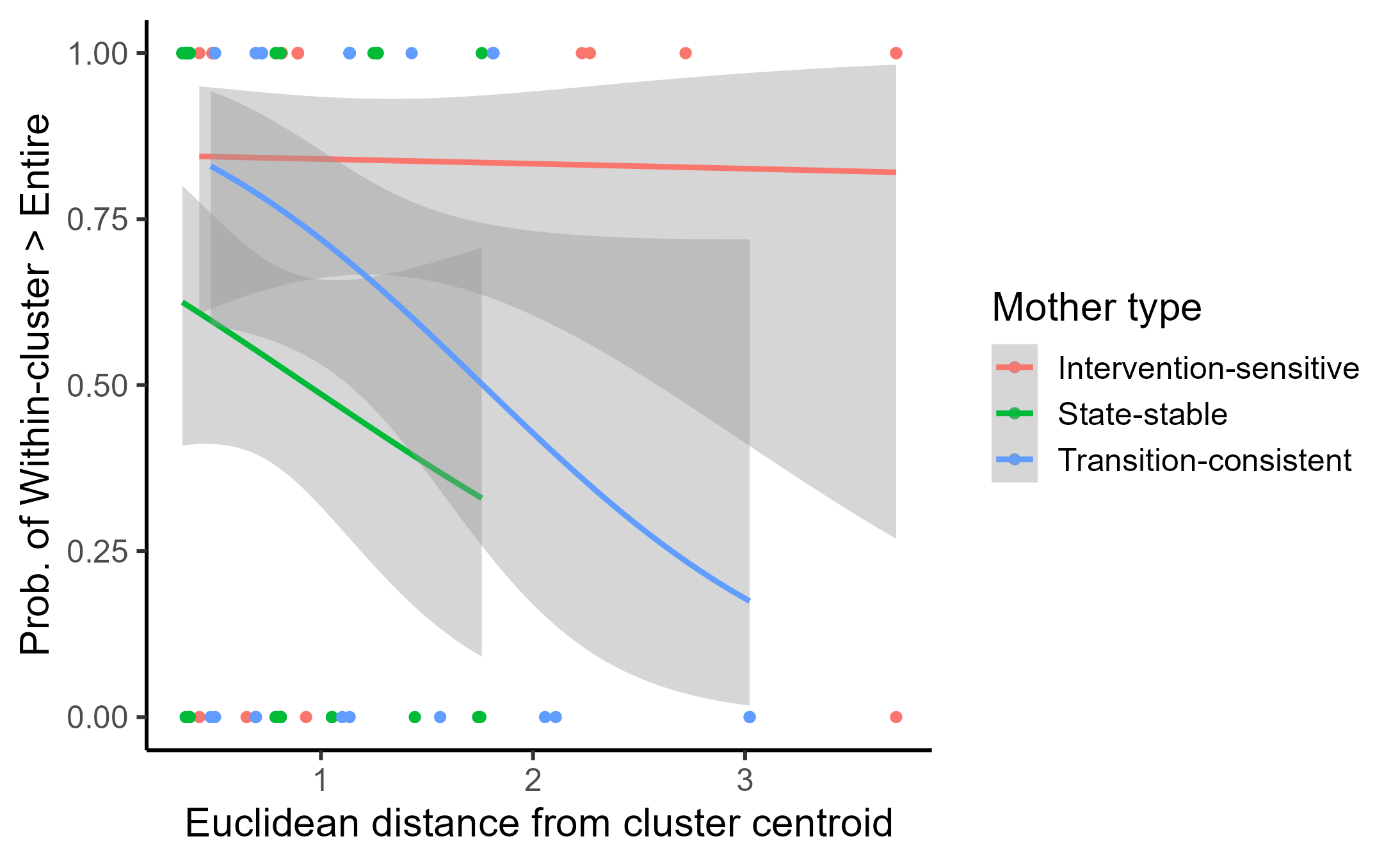}
    \caption{Probabilities of the Within-cluster vs. Entire method resulting in a lower prediction error. If a mother is better predicted by the Within-cluster method, she is assigned a probability of 1. Lines show predictions by binomial regression models, and shaded areas reflect the 95\% CIs.}
    \label{fig:pred-error-regress}
\end{figure}

Follow-up exploratory analyses with the Within-cluster method suggest that mothers closer to their cluster's centroid are better predicted by the Within-cluster method as compared to the Entire method. Three binomial regression models were constructed, one for each cluster; for a given mother, the Euclidean distance between her IBL model's attribute weights and their cluster's centroid (dist) was used to predict the method that resulted in a lower prediction error. Although none of the effects were significant due to the imbalanced and small classes, for State-stable (\(\beta_{dist}=-0.8637\)) and Transition-consistent mothers (\(\beta_{dist}=-1.235\)), the further they were from the cluster centroid, the less likely they would be better predicted by the Within-cluster method (see Figure~\ref{fig:pred-error-regress}).

\section{Conclusions and Future work}

In programs with limited resources to provide interventions to a large number of beneficiaries, a key question is to decide which subset of beneficiaries should receive an intervention. In this paper, we develop a novel approach that uses computational cognitive models based on Instance-based Learning Theory \cite{gonzalez2003instance,gonzalez2011instance} to model the human dynamic transitions of being engaged or disengaged in a program. Given the non-Markovian nature of the transitions of a beneficiary between states \cite{danassis2023limited}, a cognitive algorithm that reflects the effects of time, attention, and context similarity on human memory and decisions would simultaneously provide more accurate predictions of beneficiaries' engagement dynamics and more meaningful individual differences that other prediction methods can also leverage.

We tested our method on real-world engagement data from mothers enrolled in a maternal health program, and demonstrated that it results in higher next-step prediction accuracy compared to existing time-series methods, such as LSTM models. The personalized IBL models revealed that mothers could be clustered into three types based on their IBL attribute weights, specifically, how fluctuations in their behaviors could be attributed to intervention-sensitivity and transition-consistency. These clusters were also shown to improve the prediction accuracy of other time-series models; LSTMs trained and tested on within-cluster data outperformed all other LSTM training methods. 

Future work could explore personalizing interventions and intervention schedules based on these identified individual differences in behavioral characteristics. For example, intervention-sensitive beneficiaries could receive more frequent but shorter calls. In comparison, transition-consistent beneficiaries could receive fewer but longer calls to maximize their engagement level on each call.

One limitation of this work is the lack of an unbiased counterfactual generator to simulate the effect of recommended interventions; in our work, we used a LSTM model trained on the full dataset to generate counterfactuals. The sequence of engagement states generated by this LSTM model most likely does not reflect the true engagement dynamics of the actual mothers, and another LSTM model would better predict the generated counterfactual states due to the similarities in model architecture. This possibly explains the apparent contradiction between the IBL models outperforming the LSTM models on prediction accuracy (which does not involve counterfactuals) but underperforming on number of engaged mothers (which requires generating counterfactuals). Future work should prioritize testing our method with other datasets, particularly with a paradigm where the effects of recommended interventions on participants' task engagement can be experimentally measured, rather than relying on potentially biased counterfactuals. 

Another limitation is that our method makes predictions only from the 26th week of a mother's pregnancy onward, as the first 25 weeks are needed to train and personalize the IBL model. A possible extension of this work could involve exploring hybrid methods, such as initially relying on predictions from other trained IBL models and gradually transitioning to the personalized model.

\begin{acks}
This research was supported by the AI Research Institutes Program funded by the National Science Foundation under AI Institute for Societal Decision Making (AI-SDM), Award No. 2229881
\end{acks}


\bibliographystyle{ACM-Reference-Format}
\bibliography{references}

\appendix
\section{Real-world ARMMAN Dataset}
\label{sec:appendix_armman_results}
\subsection{Secondary Analysis}
Our experiment falls into the category of secondary analysis of the data shared by ARMMAN. 
This paper does not involve deploying the proposed algorithm or any other baselines to the service call program. As noted earlier, the experiments are secondary analysis with approval from the ARMMAN ethics board.

\subsection{Consent and Data Usage}
\label{sec:appendix_consent_data_usage}
Consent is obtained from every beneficiary enrolling in the NGO's mobile health program. The data collected through the program is owned by the NGO and only the NGO is allowed to
share data. In our experiments, we use anonymized call listenership logs to calculate empirical transition probabilities. No personally
identifiable information (PII) is available to us.
The data exchange and usage were regulated by clearly
defined exchange protocols including anonymization, read-access
only to researchers, restricted use of the data for research purposes
only, and approval by ARMMAN’s ethics review committee.

\subsection{Interventions in ARMMAN data}
The interventions in ARMMAN data are chosen based on restless multi-arm bandit (RMAB) algorithms \cite{mate2022field,zhao2024bandit}. RMABs are a model for sequentially distributing scarce resources to a set of agents \cite{whittle1988restless,xiong2023finite,jaques2019social}. 
Concretely, we have a set of arms and a limited budget and face the question of deciding which arms to pull in each round. 
The state of arms evolves according to a Markov Decision Process where transition probabilities depend on whether the arm is pulled in this step. RMABs have a broad range of applications, including resource allocation in anti-poaching, machine maintenance, cellular networks \cite{modi2019transfer,zhao2023towards}. RMABs have especially be used in healthcare settings such as call scheduling in a maternal and child care program \cite{behari2024decision,vermagroup}, screening patients at risk of cancer \cite{lee2019optimal}, and allocating hepatitis C treatment \cite{ayer2019prioritizing}.

\begin{figure}[ht!]
    \centering
    \includegraphics[width=\linewidth]{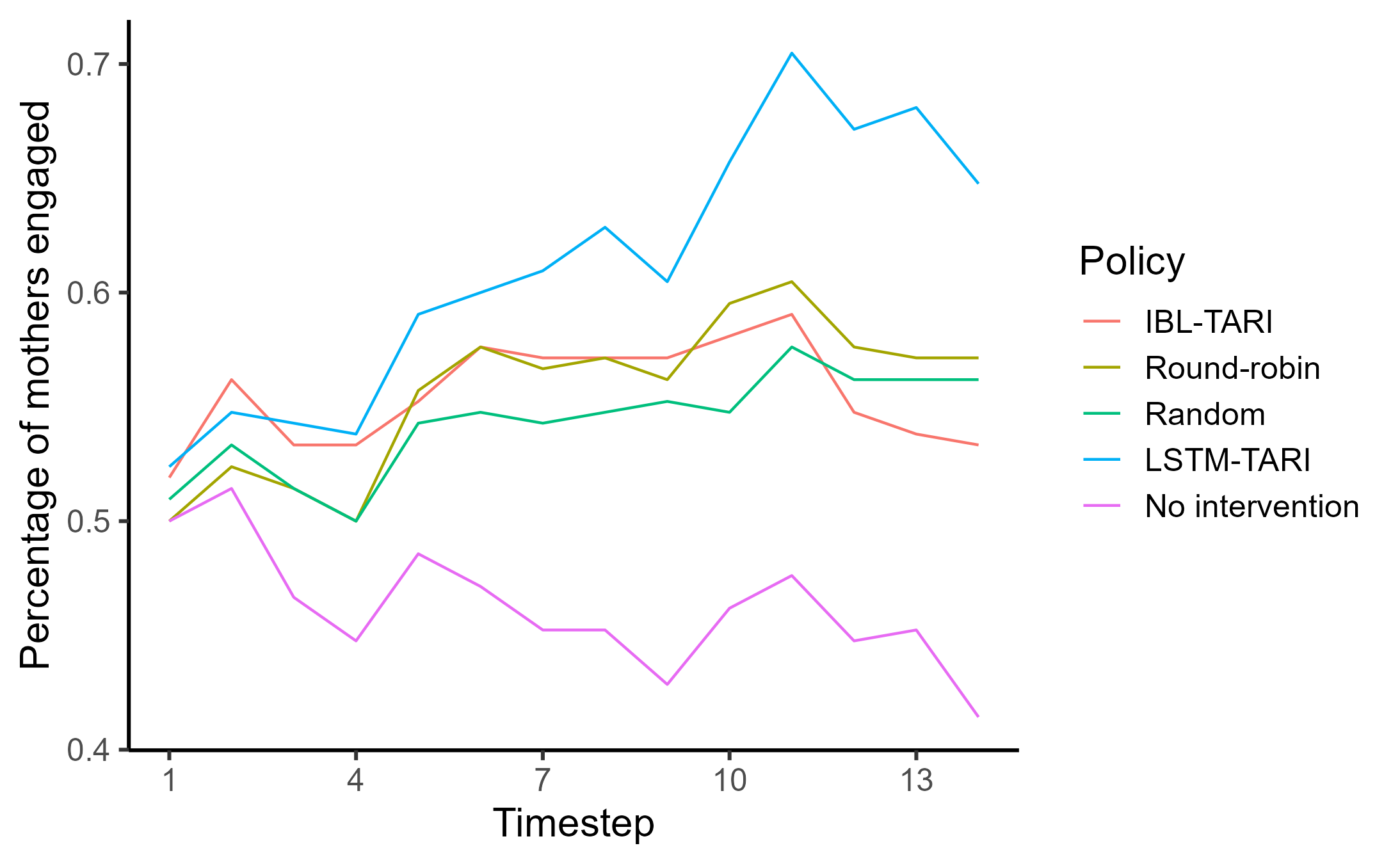}
    \caption{Percentage of mothers engaged in the program per testing time step. \(N\) = 210, budget \(k\) = 6.}
    \label{fig:percent-engaged}
\end{figure}

\end{document}